\pdfoutput=1

\documentclass[11pt]{article}

\usepackage[preprint]{acl}

\usepackage{times}
\usepackage{latexsym}

\usepackage[T1]{fontenc}

\usepackage[utf8]{inputenc}

\usepackage{microtype}

\usepackage{inconsolata}

\usepackage{graphicx}
\usepackage{multirow}
\usepackage{graphics}

\usepackage{amsmath} 
\usepackage{url}            
\usepackage{booktabs}       
\usepackage{amsfonts}       
\usepackage{nicefrac}       
\usepackage{xcolor}         
\usepackage{wrapfig}

\title{Preserving Knowledge in Large Language Model with Model-Agnostic Self-Decompression}

\author{
 \textbf{Zilun Zhang\textsuperscript{1}}$\dagger$,
 \textbf{Yutao Sun\textsuperscript{1}}$\dagger$,
 \textbf{Tiancheng Zhao\textsuperscript{2}},
 \textbf{Leigang Sha\textsuperscript{1}},
\\
 \textbf{Ruochen Xu\textsuperscript{3}},
 \textbf{Kyusong Lee\textsuperscript{2}},
 \textbf{Jianwei Yin\textsuperscript{1}},
\\
\\
 \textsuperscript{1} College of Computer Science and Technology, Zhejiang University, \\
 \textsuperscript{2}Binjiang Research Institute of Zhejiang University, \\
 \textsuperscript{3}Linker Technology Research Co. Ltd
\\
 \small{
   \textbf{Correspondence:} \href{tianchez@zju-bj.com}{tianchez@zju-bj.com}
 }
$\dagger$: Equal Contribution
}

\begin{document}
\maketitle
\begin{abstract}

Humans can retain old knowledge while learning new information, but Large Language Models (LLMs) often suffer from catastrophic forgetting when post-pretrained or supervised fine-tuned (SFT) on domain-specific data. 
Moreover, for Multimodal Large Language Models (MLLMs) which are composed of the LLM base and visual projector (e.g. LLaVA), a significant decline in performance on language benchmarks was observed compared to their single-modality counterparts.
To address these challenges, we introduce a novel model-agnostic self-decompression method, \textbf{Tree Generation (TG)}, that decompresses knowledge within LLMs into the training corpus. This paper focuses on TG-SFT, which can synthetically generate SFT data for the instruction tuning steps. 
By incorporating the dumped corpus during SFT for MLLMs, we significantly reduce the forgetting problem. 

\end{abstract}

\section{Introduction}


The Large Language Models (LLMs) and Multimodal Large Language Models (MLLMs) have been rapidly developed and iterated in recent years. Many of them show a significant leap in the capability of understanding, generation, and interaction following the natural language \cite{openai2024gpt4, geminiteam2024gemini, anthropic2024claude}. There are lots of LLMs and MLLMs that have been developed in practice \cite{kaddour2023challenges, yin2024survey}. However, the model trained for the general purpose may have a decline in performance in specific domains such as math, coding, law, healthcare, finance, etc. \cite{wu2024llama}, therefore the need for obtaining sufficient training data to develop domain-specific LLMs or MLLMs is crucial.

Collecting extensive domain-specific data and training LMMs from scratch is challenging. As a result, post-pretraining \cite{gururangan2020dont} or supervised fine-tuning \cite{brown2020language} (SFT) of general LLMs/MLLMs with domain-specific data become the popular strategy for those seeking domain-specific models \cite{codellama, huang2023lawyer, azerbayev2024llemma, yunxiang2023chatdoctor, li2023llavamed, kuckreja2023geochat}. 
However, this process can impair the models' performance due to catastrophic forgetting \cite{aleixo2023catastrophic, luo2024empirical}. 
We need the expert model to be generalizable as the general model on the specific domain.
Although Parameter Efficient Finetuning (PEFT) methods \cite{peft} can adapt the models to the new domain by adding only a few parameters and maintaining their original capabilities, they often result in less satisfactory performance and are hard to accumulate from different domains.  This calls for an approach that integrates domain-specific expertise into LLMs/MLLMs without compromising their general capabilities.



\begin{figure}[t]
  \includegraphics[width=\columnwidth]{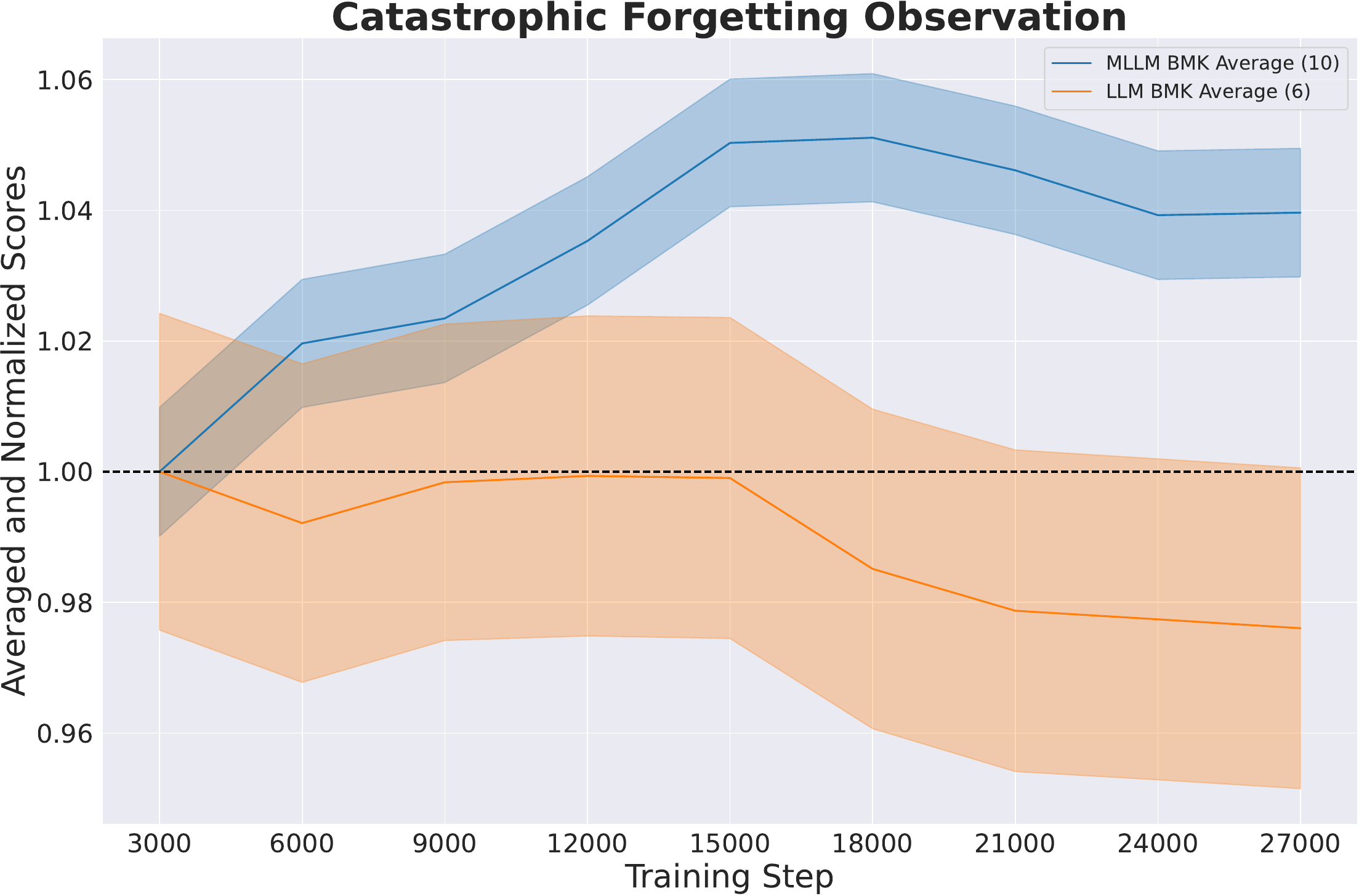}
  \caption{The motivation of Our Work. Shadow represents the error bar. The SFT of MLLM harms the language ability of its LLM backbone (MLLM has begun to forget its general language ability while training is processed). We choose the LLaMA2-7B-chat model as the LLM backbone for the experiments. Details of this experiment can be found in Appendix \ref{appendix:figure1}. The first data point is evaluated from the checkpoint of 3000 steps. We averaged the results of 10 MLLM benchmarks and 6 LLM benchmarks respectively and normalized them with the result of the first checkpoint to show the trend (Increased performance if the score is greater than one. Decreased performance if the score is less than 1).
  }
  \label{fig:motivation}
\end{figure}

The problem of catastrophic forgetting in LLM is widely discussed. To verify the problem of catastrophic forgetting in MLLM, we trained (SFT) LLaVA \cite{liu2023llava} for 5 epochs, and evaluated the model every 3000 steps. As shown in Figure \ref{fig:motivation}, we observed that the performance of MLLM benchmarks grew even after the third epoch, but LMM benchmarks started to deteriorate since the third epoch. These different behaviors of the model performance between vision-language benchmarks and pure language benchmarks indicate that MLLM has begun to forget its general language ability rather than simply overfit the data.

There is a point of view that describes the LLMs as lossless compressors \cite{jack_rae_compression, ilya, gu2024optimal, yang2023introduction, chatgpt_blurry_jpeg_Web}. \cite{delétang2024language} from DeepMind explores the connection between predictive models and lossless compression and state that the advancement in self-supervised LLMs can be effectively leveraged for compression tasks. They demonstrate that LLMs not only excel in text compression but also show competitive performance across different data modalities, such as images and audio.

Enlightened by these works, we considered the process of synthesizing data with LLMs (a.k.a. generating text data) from LLMs as a decompression process. We aim to preserve knowledge from LLMs by taking a snapshot of the LLMs, i.e. using LLMs as offline data generators and dumping the generated corpus. This self-decompression method should be able to apply to any LMM (model-agnostic). By adding the decompressed data during post-pretraining or SFT, the old knowledge could be reminded and kept. For this purpose, we design a novel approach, named Tree-Generation (\textbf{TG}), along with its variants TG-SFT for supervised fine-tuning the MLLMs. With this model-agnostic approach, we observed the catastrophic forgetting problem can be reduced significantly. 

From extensive experiments, we show that \textbf{TG} algorithm is useful in reducing catastrophic forgetting. Our contribution on \textbf{TG} algorithm can be summarized as threefolds. 

\begin{itemize}
    \item \textbf{TG} algorithm is a self-contained data generation algorithm based on LLMs, rather than for any specific NLP task (i.e. training BERT includes many specific NLP tasks).
    \item \textbf{TG} algorithm is universally applicable to any LLMs for SFT (TG-SFT).
    Importantly, no additional manual prompt is required.
    \item \textbf{TG} algorithm is quite foundational, hence it has many applications such as preventing catastrophic forgetting, knowledge distillation, continual learning, etc.
\end{itemize}

\section{Related Work}


\subsection{Methods for Preventing Catastrophic Forgetting}

Preventing Catastrophic Forgetting during training is a classic topic in deep learning. Since 2018, many works discussed how to migrate the Catastrophic Forgetting for LLMs. \cite{yang2024selfdistillation} introduced a method that uses self-distillation to bridge the distribution gap between task datasets and LMMs, mitigating catastrophic forgetting while preserving general capabilities. \cite{luo2024empirical} conducted an empirical investigation revealing the prevalence of catastrophic forgetting in LLMs as model scale increases during continual instruction tuning, with suggestions that general instruction tuning can help alleviate this issue. \cite{hsieh2023distilling} designed a method for training smaller models that outperform LLMs with less training data by extracting rationales from LLMs as additional supervision.
Furthermore, \cite{dou2024loramoe} and \cite{wang2023orthogonal} demonstrate LoRAMoE and O-LoRA, the former introduces low-rank adapters and a router network to alleviate world knowledge, and the latter mitigate catastrophic forgetting by learning new tasks in orthogonal subspaces to minimize interference with past knowledge was proposed by \cite{wang2023orthogonal}.

\begin{figure*}[t]
  \includegraphics[width=\linewidth]{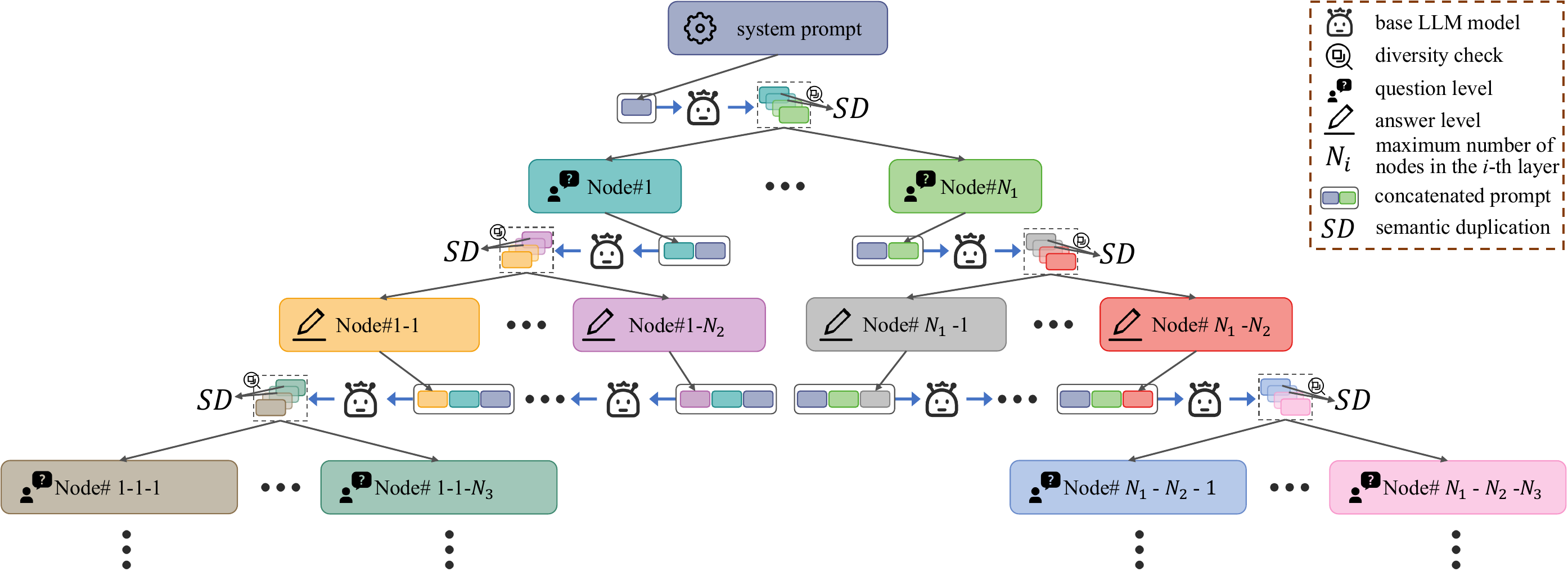}
  \caption{TG-SFT structure overview, illustrates a three-layer complete tree structure. In practice, the depth of different leaf nodes can be adjusted as needed.  This figure depicts a typical form of a Balance-Tree, whereas in a Wide-Tree, no further branching occurs beyond the second layer. Starting from the first layer, all odd-numbered layers serve as question layers, and even-numbered layers serve as answer layers.}
  \label{fig:tg_sft}
\end{figure*}

\subsection{Sythetic Data for LLM Training}
LLMs are trained or tuned on vast amounts of data. Due to the data shortage (especially high-quality data) in many specialized domains, the synthesis of training data for LLMs is crucial. 

\cite{liu2024best} provide a comprehensive review of the use of synthetic data in training and evaluating AI models, highlighting its potential to overcome data scarcity while emphasizing the importance of ensuring data quality for responsible AI development.  \cite{yehudai2024genie} present Genie, a method for the automatic generation of high-quality, content-grounded datasets through a three-stage process: content preparation, generation, and filtering. Researchers from Microsoft introduce 1.3B models  \cite{gunasekar2023textbooks, li2023textbooks} that achieve competitive performance on code generation tasks by leveraging a novel training approach using 'textbook quality' synthetic data, demonstrating the potential for smaller models to rival larger ones through high-quality data curation. \cite{xu2023wizardlm} present Evol-Instruct, an evolutionary algorithm that generates diverse and complex instruction data, enhancing LLM performance on high-complexity tasks through automated data evolution and fine-tuning. \cite{mitra2024orcamath} introduce Orca-Math, which utilizes a high-quality synthetic dataset of 200K math problems and an iterative learning technique. \citet{li2023synthetic} find a negative correlation between the models' performance when trained on synthetic data and the degree of subjectivity involved in classification tasks. MAGPIE was presented by \cite{xu2024magpie}, a method for synthesizing high-quality instruction data from aligned LLMs by prompting them with minimal input. Nvidia released Nemotron-4 340B \cite{nvidiaNVIDIAReleases}, a model trained on 9 trillion tokens, and over 98\% of the data used in the model alignment process was synthetically generated.


\subsection{Data Extraction from LLMs and LLM Self-Iteration}

Jang explores the capability of GPT-4 to self-reflect and edit its own generations, suggesting the potential for self-correction and improvement in LLMs without external feedback \cite{jang2023reflection}. Lee et al. \cite{lee2024llm2llm} introduce a targeted and iterative data augmentation strategy that enhances the performance of LLMs in low-data regimes by using a teacher LLM to generate synthetic data based on incorrect predictions from a student model. Finlayson et al. \cite{finlayson2024logits} demonstrate that non-public information about API-protected LLMs can be gleaned from a small number of API queries, due to the softmax bottleneck in LLM architectures, with implications for model transparency and accountability. Nasr et al. \cite{nasr2023scalable} presents a study on the extractable memorization in language models, showing that adversaries can efficiently extract significant amounts of training data from various models, including open-source, semi-open, and closed models, highlighting the need for improved privacy protections.

\section{Methodology}


TG-SFT is a method designed for high-quality, efficient dialogue generation using a backbone LLM. This approach builds structured dialogue sequences through a tree-based expansion strategy, which is shown in Figure \ref{fig:tg_sft}, aiming to produce diverse and accurate conversational corpora for model training.
\subsection{Initialization and Layered Expansion}

The TG-SFT methodology initiates with a general system prompt, denoted as $P_0$, which is augmented by a special marker indicating the start of instruction. This composite prompt serves as the input to the backbone LLM, prompting it to generate various questions as if it were in the user's role. Formally, we express the first layer input as:
$$P_1 = P_0 + \text{"<user>"}$$

where $\text{"<user>"}$ signifies the special marker for instructional onset. Based on $P_1$, the backbone LLM produces a set of $N_1$ (which is the predefined number of nodes for the first layer) questions, which after semantic deduplication (\textit{SD}), form the child nodes of the first layer. We use MMR (Maximal Marginal Relevance) \cite{DBLP:journals/sigir/CarbinellG17} algorithm to conduct semantic deduplication process. MMR  is a technique used to enhance diversity in generated text by balancing relevance and novelty. To apply MMR effectively, we use SentenceBERT \cite{DBLP:conf/emnlp/ReimersG19} to convert text into vectors. This allows the algorithm to evaluate both the relevance of each sentence to the topic and its distinctiveness from previously selected sentences, thereby reducing redundancy and enhancing the informativeness of the output.

\subsection{Recursive Dialogue Generation}
For each question generated at node \( i \) in the first layer, the formulation of the prompt for the subsequent layers involves dynamically appending the sequence of alternating roles of $\text{"<user>"}$ and $\text{"<assistant>"}$. Each layer's prompt is constructed by appending the relevant question or response to the initial prompt \( P_0 \), supplemented by role markers to guide the model's generation contextually.

For odd-indexed layers (\( 2i+1, i \geq 0\)), representing user-initiated questions, the prompt is constructed as follows:
\begin{align*}
    P_{2i+1} = & P_0 \hspace{0.5mm} + \text{"<user>"} + Q_1 + \text{"<assistant>"} +\\
    &  R_1 + \ldots + R_i + \text{"<user>"}
\end{align*}
Here, \( Q_1, R_1, \ldots, R_i \) represent the sequence of questions and responses up to the \( i \)-th response, with the subsequent user query being formulated. This setup directs the base LLM to continue the dialogue from the perspective of the user, generating a new question \( Q_{i+1} \).

\begin{figure}[t]
  \includegraphics[width=\linewidth]{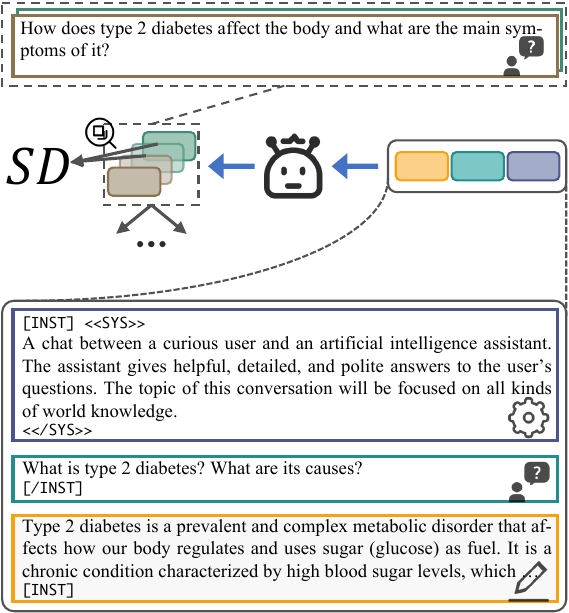}
  \caption{Example of Concatenated Prompts: This figure uses the Llama2-chat model as the backbone LLM. In Llama2, the system prompt is enclosed with \text{"<<SYS>>"}, \text{"[INST]"} indicates the start of an instruction, signifying the beginning of generation in the user role. \text{"[/INST]"} marks the end of the instruction. LLMs are trained to start responding from this point in the pre-training phase.}
  \label{fig:prompt_example}
\end{figure}

For even-indexed layers (\( 2i \)), which capture responses from the assistant, the prompt configuration is:
\begin{align*}
    P_{2i} = & P_0 \hspace{0.5mm}+ \text{"<user>"} + Q_1 + \text{"<assistant>"} +\\
    &  R_1 + \ldots + Q_i + \text{"<assistant>"}
\end{align*}
In this case, \( Q_1, R_1, \ldots, Q_i \) denote the alternating questions and responses leading to the \( i \)-th question, setting the stage for the assistant's response. Examples of $P_2i$ and $P_{2i+1}$ are illustrated in the Figure \ref{fig:prompt_example}

\subsection{Corpus Construction}
The final structure comprises nodes at depth $2k$, with a total of $\prod_{i=1}^{2k} N_i$ leaf nodes. From the root $P_0$, any path leading to a leaf node at depth $2k$ represents a complete dialogue sequence. In the TG-SFT model, we use $C$ to represents the set of all possible dialogue sequences generated from the root to the leaves of the tree. Formally, $C$ is defined as:
$$C = \bigcup_{i=1}^{N_{2k}} \left( P_0 + Q_1 + R_1 + \ldots + Q_k + R_k \right)_i$$
where $Q_j$ and  $R_j$ represent the questions and responses at each dialogue turn $j$.
\subsection{Structural Features of TG-SFT}
The TG-SFT algorithm is strategically designed to optimize dialogue generation by varying the breadth and depth of the generated dialogue tree in response to the complexity of the conversation.

\textbf{Initial Question Generation Layer:}
The first layer $N_1$ is typically larger, allowing the model significant latitude to explore various topics arising from the general system prompt \( P_0 \). This expansive approach is crucial as it branches out into \( N_1 \) distinct questions, laying a broad thematic groundwork.

\textbf{Specificity and Depth in Subsequent Layers:}
The sizes of subsequent layers are tailored based on the specific requirements of the dialogue depth. For any \( q \) (where \( q \geq 0 \)), \( N_{2q+1} \) represents the number of new questions posed in response to the answers at layer \( 2q \). As discussions become more specific, the potential to branch out further diminishes, generally resulting in \( N_{2p+1} < N_{2q+1} \) for \( p > q \). Similarly, \( N_{2q+2} \) represents the number of responses to the questions at layer \( 2q+1 \), with \( N_{2p+2} \) typically being less than \( N_{2q+2} \) as the conversation narrows down.

\textbf{Token Allocation Strategy:}
The token allocation per layer, \( L_i \), is carefully designed. For layers \( 2q+1 \), a constant \( m_0 \) is set, representing the maximum token count for any question, ensuring questions remain concise while enhancing the efficiency of the generation process:
$$L_{2q+1} = m_0, \forall q$$
For layers that involve responses (\( 2q+2 \)), \( L_{2q+2} \) is typically smaller than \( L_{2p+2} \) for deeper layers, reflecting the increased specificity and detail of responses as the dialogue progresses:
$$L_{2q+2} < L_{2p+2}, \text{ for } p > q$$
This setting aims to provide more detailed explanations as the discussion delves deeper into some specific topics.

\textbf{Flexibility and Customization:}
Both \( N_i \) and \( L_i \) are tunable parameters, offering the flexibility to enrich the dialogue corpus significantly. By setting a higher \( N_1 \), the algorithm ensures a wide range of initial topics, leading to a rich and diverse corpus. This diversity, coupled with detailed responses in deeper layers, ensures comprehensive coverage of specific topics, providing detailed and contextually relevant answers. There is a special type of TG-SFT structure called \textbf{Wide-Tree} where \( N_1 \) is set exceptionally high while \( N_i \) (for \( i > 1 \)) are all equal to 1. Conversely, trees where \( N_i \) values (for \( i > 1 \)) are not all set to 1 are referred to as \textbf{Balance-Tree}, which allow for a less extensive initial exploration yet more detailed follow-up inquiries across the subsequent layers. We will further discuss TG-SFT(Wide-Tree) and TG-SFT(Balance-Tree) in section \ref{wide-tree}.

\section{Experiments}
In this section, we introduce our experiment settings and discuss the findings. 
We begin this section by detailing the experiment configuration (section \ref{expsection:config}). 
Then we validate the effectiveness of the TG-SFT approach (section \ref{expsection:mainresult}). 
Next, we demonstrate the potential of TG-PT (section \ref{expsection:tgpt}), the application of \textbf{TG} in post-pretraining. Subsequently, we analyze the corpus generated by TG-SFT (section \ref{expsection:dataanalysis}). Finally, in section \ref{expsection:treeconfig} and \ref{expsection:numturns}, we explore the impact of different tree configurations and number of conversation turns.

\subsection{Experimental Settings}

\paragraph{Models.} Unless specified, we use the LLaMA2-chat (7B) model \cite{touvron2023llama} for the experiments in this section. For the MLLM, we train a projector to align the CLIP vision encoder with LLaMA2-chat. Then, we supervisedly fine-tune (SFT) both the LLaMA2-chat model and the projector to obtain the LLaVA model \cite{liu2023llava}. We choose the LLaMA2-chat model as the LLM backbone to avoid the influence of additional tuning data, such as ShareGPT used for Vicuna. All LLaVA models discussed in this paper use LLaMA2-chat as the LLM backbone. The codebase for alignment and SFT is obtained from the official LLaVA GitHub repository \footnote{https://github.com/haotian-liu/LLaVA}.


\paragraph{Data.} In line with LLaVA, we use the 558K subset of the LAION-CC-SBU dataset \footnote{https://huggingface.co/datasets/liuhaotian/LLaVA-Pretrain} to align the vision encoder and the LLM. For supervised fine-tuning, we employ the Mix 665K dataset \footnote{https://huggingface.co/datasets/liuhaotian/LLaVA-Instruct-150K/blob/main/llava\_v1\_5\_mix665k.json}. Additionally, we generate a 100K language-only corpus using the \textbf{TG} approach.


\paragraph{Evaluation.} For MLLM benchmarks, we follow LLaVA and select the following datasets: GQA~\cite{gqa}, MMBench~\cite{mmbench}, POPE~\cite{pope}, ScienceQA-IMG~\cite{sciqaimage}, SEED-Bench~\cite{seedbench}, TextVQA~\cite{textvqa}, VisWiz~\cite{vizwiz}, and VQA$^{v2}$~\cite{vqav2}. For LLM benchmarks, we adhere to LLaMAPro \cite{wu2024llama} and choose the AI2 Reasoning Challenge (ARC, 25-shot) ~\cite{clark2018think}, HellaSwag (10-shot)~\cite{zellers2019hellaswag}, MMLU (5-shot)~\cite{hendrycks2021measuring}, TruthfulQA (0-shot)~\cite{lin2022truthfulqa}, and Winogrande (5-shot)~\cite{sakaguchi2019winogrande}. We use lmms-eval \cite{lmms_eval2024} as the MLLM evaluation pipeline and lm-evaluation-harness \footnote{https://github.com/hills-code/lm-evaluation-harness} as the LLM evaluation pipeline, as proposed by Gao et al. \cite{eval-harness}. Additionally, we report the average scores for both MLLM and LLM benchmarks.


\paragraph{Training Details.} Our experiments are conducted on 8 A100-80GB GPUs with NVLink. As observed in Figure \ref{fig:motivation}, SFT causes the model to start forgetting LLM knowledge by the third epoch. Therefore, we set the number of training epochs for SFT to 3, which requires 36 hours of training. The batch size is set to 32 per device, and other training parameters follow LLaVA's defaults. The corpus generation process takes between 20 to 40 hours on 8 A100-40GB GPUs without NVLink, depending on the configuration of the TG-SFT approach.

\label{expsection:config}
\subsection{Results of TG-SFT}
\label{expsection:mainresult}

\begin{table*}[ht]
\centering  
\scalebox{0.9}{
\begin{tabular}{l|cccccccc|c}  
\toprule
\multirow{2}{*}{Method Name}   & \multicolumn{8}{c|}{\textbf{MLLM Benchmark}} &  \multirow{2}{*}{Average} \\
     & GQA   & MMB   & POPE   & SQA$^\text{I}$   & SEED & VQA$^T$ & VisWiz & VQA$^{v2}$ &  \\ 
    \midrule
  \multicolumn{8}{l}{\it Model-wise Result (Baseline \& Model Augmented) } \\   
 LLaMA2-chat (LLM) & - & - & - & - & - & - & - & - & -  \\
 LLaVA (Full-Param) & 62.55	& 63.66	&	85.71	&69.31	&66.08&	45.28&	54.79&	77.19 & 65.57 \\
LLaVA (LoRA) & 63.16 &	64.26&		85.32&	66.78&	66.41&	46.26&	52.36&	77.71& 65.28\\
\midrule
\multicolumn{8}{l}{\it Data-wise Result (Data Augmented)} \\   
Human (ShareGPT) & 62.64	&63.57	&	84.63	&67.28&	65.33&	44.89&	53.61	&77.02	& 64.87 \\ \hline
TG-SFT (Wide-Tree) & 62.26 & 64.60 & 84.69 & 68.47 & 65.53 & 45.35 & 52.41 & 77.04 & 65.04  \\
TG-SFT (Balance-Tree) & 62.79 &	64.35&		85.13&	68.02&	65.11&	45.44&	52.45&	77.02 & 65.04  \\
\bottomrule
\end{tabular}
}
\scalebox{0.9}{
\begin{tabular}{l|cccccc|c}  
\toprule
\multirow{2}{*}{Method Name}   & \multicolumn{6}{c|}{\textbf{LLM Benchmark}} &  \multirow{2}{*}{Average} \\
      & ARC & HellaSwag & MMLU & TruthfulQA & Winogrande & GSM8K & \\ 
    \midrule
  \multicolumn{8}{l}{\it Model-wise Result (Baseline \& Model Augmented) } \\   
 LLaMA2-chat (LLM) & 53.50	& 78.58 &	47.24&	45.32	& 72.53&23.28  &  53.41  \\
 LLaVA (Full-Param) & 49.06	& 72.71&	47.98&	49.26&	68.75&	15.85 & 50.60 \\
LLaVA (LoRA) & 51.02&	74.08&	48.75	&47.49&	70.24	&16.98 & 51.43\\
\midrule
\multicolumn{8}{l}{\it Data-wise Result (Data Augmented)} \\   
Human (ShareGPT) & 54.69	&74.05	& 50.60	& 49.86 &71.90  &	21.38 & 53.75 \\ \hline
TG-SFT (Wide-Tree) & 49.06 & 76.48 & 50.19 & 50.30 & 70.48 & 21.30 & 52.97 \\
TG-SFT (Balance-Tree) & 53.41	& 75.83&	50.09&	50.69	&70.01&	20.77 & \textbf{53.47} \\
\bottomrule
\end{tabular}
}
\vspace{1mm}
\caption{
\textbf{
Comparison with different approaches on 8 MMLM benchmarks and 6 LMM benchmarks.} 
Benchmark names are abbreviated. MMB: MMBench; SQA$^\text{I}$: ScienceQA-IMG; SEED: SEED-Bench; VQA$^\text{T}$: TextVQA; VQA$^{v2}$: VQA-v2  ; ARC: AI2 Reasoning Challenge. Compared with the LLaVA baseline, 
LLaVA trained with the Wide-Tree TG-SFT approach restores the average score of the LLM benchmark from 50.60 to 52.97. TG-SFT (Balance-Tree) further boosts this performance to 53.47, which is slightly higher than the LLaMA2-chat backbone's performance. TG-SFT approaches maintain a comparable performance with LLaVA (Full-Param tuned) baseline on the MLLM benchmarks as well.
}

\label{tab:main_table}  
\end{table*}


We evaluate the results of LLaVA trained with the corpus generated by TG-SFT against other approaches, as shown in Table \ref{tab:main_table}. We categorize the different approaches into two groups: Model-wise and Data-wise. All corpora in this subsection are generated using the LLaMA2-chat (7B) model, with a total of 100K conversations generated. The different approaches are explained below.


\paragraph{LLaMA2-chat.} This is the baseline for LLM benchmarks without any additional modifications.


\paragraph{LLaVA (Full-Param).} The baseline approach for MLLM benchmarks, where the LLM backbone is replaced by LLaMA2-chat. This is done because we use LLaMA2-chat to generate synthetic data in other experiments. The corresponding projector is aligned using the LAION-CC-SBU dataset as described in the original LLaVA paper.


\paragraph{LLaVA (LoRA).} The LLaVA is fine-tuned (SFT) with the Mix 665K data and LoRA \cite{hu2021lora}. During the evaluation of LLM benchmarks, the trained LoRA adapter is deactivated.


\paragraph{Human (ShareGPT).} The LLaVA is fine-tuned with the Mix 665K data and 100K data randomly selected from the ShareGPT dataset used in LLaMAPro \cite{wu2024llama}. The 100K additional training data in this approach is of high quality since it was generated and rated by humans. This approach serves as an upper bound for Data-wise approaches.

\paragraph{TG-SFT (Wide-Tree).}\label{wide-tree} The LLaVA is supervisedly fine-tuned with the Mix 665K data and 100K synthetic data randomly generated by LLaMA2-chat. This approach serves as the baseline for Data-wise approaches. This structure maximizes the breadth of initial topic exploration at the first layer, prioritizing a vast range of questions without focusing on detailed follow-up inquiries in deeper layers. Such a configuration allows for the broadest possible survey of topics, albeit at the expense of depth in specific issue elaboration.





\paragraph{TG-SFT (Balance-Tree).} The LLaVA is fine-tuned with the Mix 665K data and 100K synthetic data generated by the TG-SFT (Balance-Tree) approach. The \textbf{knowledge-guided technique} was applied, ensuring that the synthetic data follows the knowledge-guided distribution.


\paragraph{Result Analysis.} Table \ref{tab:main_table} highlights the forgetting phenomenon in MLLM during SFT. The average score for the LLM benchmark of the LLaMA2-chat model is 53.41, but the average score for the LLaVA baseline decreases to 50.60. Meanwhile, LLaVA trained with the TG-SFT (Balance-Tree) approach and Knowledge-Guided technique restores the average LLM benchmark score to 53.47 while maintaining comparable performance on the MLLM benchmarks. The average performance in the LLM benchmark for TG-SFT(Balance-Tree) nearly matches that of LLaVA SFT with human-produced ShareGPT data, demonstrating that synthetic data generated using TG-SFT is highly effective, even compared to real data.

Additionally, while the TG-SFT (Wide-Tree) approach performs better than the LLaVA (Full-Param) baseline, it does not achieve the same level of performance as LLaVA with the Knowledge-Guided TG-SFT (Balance-Tree) approach, indicating the importance of knowledge guidance.

In practice, we found that TG-SFT (Wide-Tree) is faster than TG-SFT (Balance-Tree) in terms of the data generation speed since the former can generate data in a fully parallel manner in theory, as long as we have enough VRAM. The latter one can only generate data in a partially parallel way since prior knowledge from previous layers is needed.

\subsection{TG-PT}
\label{expsection:tgpt}
Dialogue data generated by TG-SFT is suitable for SFT training. To validate the efficacy of the TG method, we have developed a new variant of TG-SFT specifically designed for Post-Pretraining. This variant involves substituting the backbone LLM from a chat model to a base model, thereby eliminating role-switching in generation. Instead, this variant initiates with a simple prompt, such as \text{"}\textit{Here are some useful world knowledge:}\text{"} and continues to expand the dialogue in a tree-structured manner. The resulting data is suitable for post-training applications. Consequently, we refer to this variant as TG-PT (Tree-Generation for Post-PreTraining).



\begin{table*}[ht]
\centering  
\scalebox{0.9}{
\begin{tabular}{l|cccccc|c}  
\toprule
\multirow{2}{*}{Method Name}   & \multicolumn{6}{c|}{\textbf{LLM Benchmark}} &  \multirow{2}{*}{Average} \\
      & ARC & HellaSwag & MMLU & TruthfulQA & Winogrande & GSM8K & \\ 
    \midrule 
LLaMA-2-Base & 54.01	& 78.63	& 45.6	&38.92	&73.95&	13.27  &  50.73  \\
Random & 51.62	& 62.07	&44.79	&41.03	&71.43	&0.45 & 45.23 \\
TG-PT & 55.55	&78.69	&45.04	&38.44	&73.56	&11.9 & 50.53\\
\bottomrule
\end{tabular}
}
\vspace{1mm}
\caption{
Result of TG-PT Algorithm in LLM Benchmarks.
}
\label{tab:tgpt_table}  
\end{table*}


Table \ref{tab:tgpt_table} demonstrates the effectiveness of the TG-PT approach. We conducted post-pretraining on the LLaMA-2 base model (7B) with 100K data. The LLM benchmark performance using data generated by TG-PT is compared to that using randomly generated data. The results indicate that post-pretraining the LLM with a randomly generated corpus leads to significant performance degradation. However, post-pretraining with data generated using the TG-PT approach not only mitigates this issue but also provides a slight performance gain. This result illustrates that the TG method is not only suitable for SFT but also for Post-Pretraining, demonstrating its versatility. This indirectly confirms the general applicability of the TG algorithm in decompressing LLMs, showcasing its efficacy across different training regimes.

\subsection{Decompressed Data Analysis}
\label{expsection:dataanalysis}

\begin{figure}[ht]
  \includegraphics[width=\columnwidth]{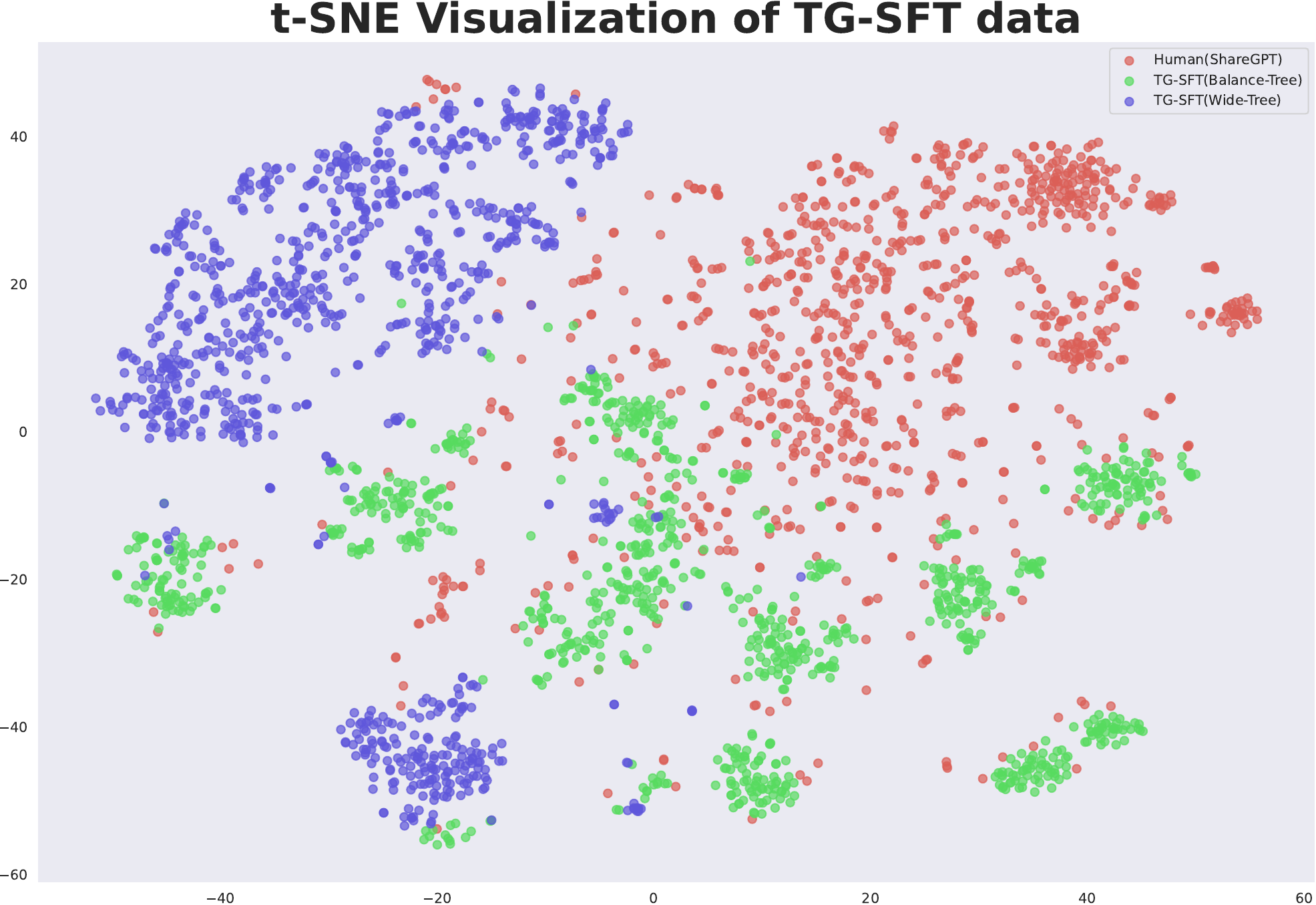}
  \caption{T-SNE data visualization for corpus generated using TG-SFT and collected from ShareGPT}
  \label{fig:decompressed_data}
\end{figure}

As shown in Figure \ref{fig:decompressed_data}, we randomly sampled 1K conversations from three different sources: data generated with TG-SFT (Wide-Tree), TG-SFT (Balance-Tree), and data collected from ShareGPT. We then visualized these samples using SentenceBert embeddings ("all-MiniLM-L6-v2" model) \cite{reimers-2020-multilingual-sentence-bert} and T-SNE \cite{JMLR:v9:vandermaaten08a}. The clusters for these three types of corpus are quite distinctive.

\subsection{Tree Configuration}
\label{expsection:treeconfig}
\begin{table}[ht]
\centering
\scalebox{0.7}{
\begin{tabular}{c|cccc|cc}
\toprule
Exp ID& L1 (Q) & L2 (A) & L3 (Q) & L4 (A) & MLLM & LLM \\
\midrule
1 & 32 & 16 & 8 & 8 & 65.04 &53.47\\
2 & 16 & 16 & 8 & 8 & 65.08&53.46\\
3 & 32 & 8 & 8 & 8 &65.21 &\textbf{53.80}\\
4 & 32 & 16 & 8 & 4 & \textbf{65.23}&52.88 \\
\bottomrule
\end{tabular}
}
\vspace{1mm}
\caption{Results of Different Tree Configurations. "L" represents "Layer", "Q" represents "Question", and "A" represents "Answer".}
\label{table:tree_config}
\end{table}


Table \ref{table:tree_config} demonstrates the performance of the TG-SFT approach with various tree configurations. Specifically, we halved the branching factor at different tree levels and generated the corpus for training. The average scores of MLLM and LLM benchmarks across different settings do not show significant differences, indicating the parameter-insensitivity property of our TG-SFT approach.

\subsection{Conversation Turns}
\label{expsection:numturns}
\begin{figure}[ht]
  \includegraphics[width=\columnwidth]{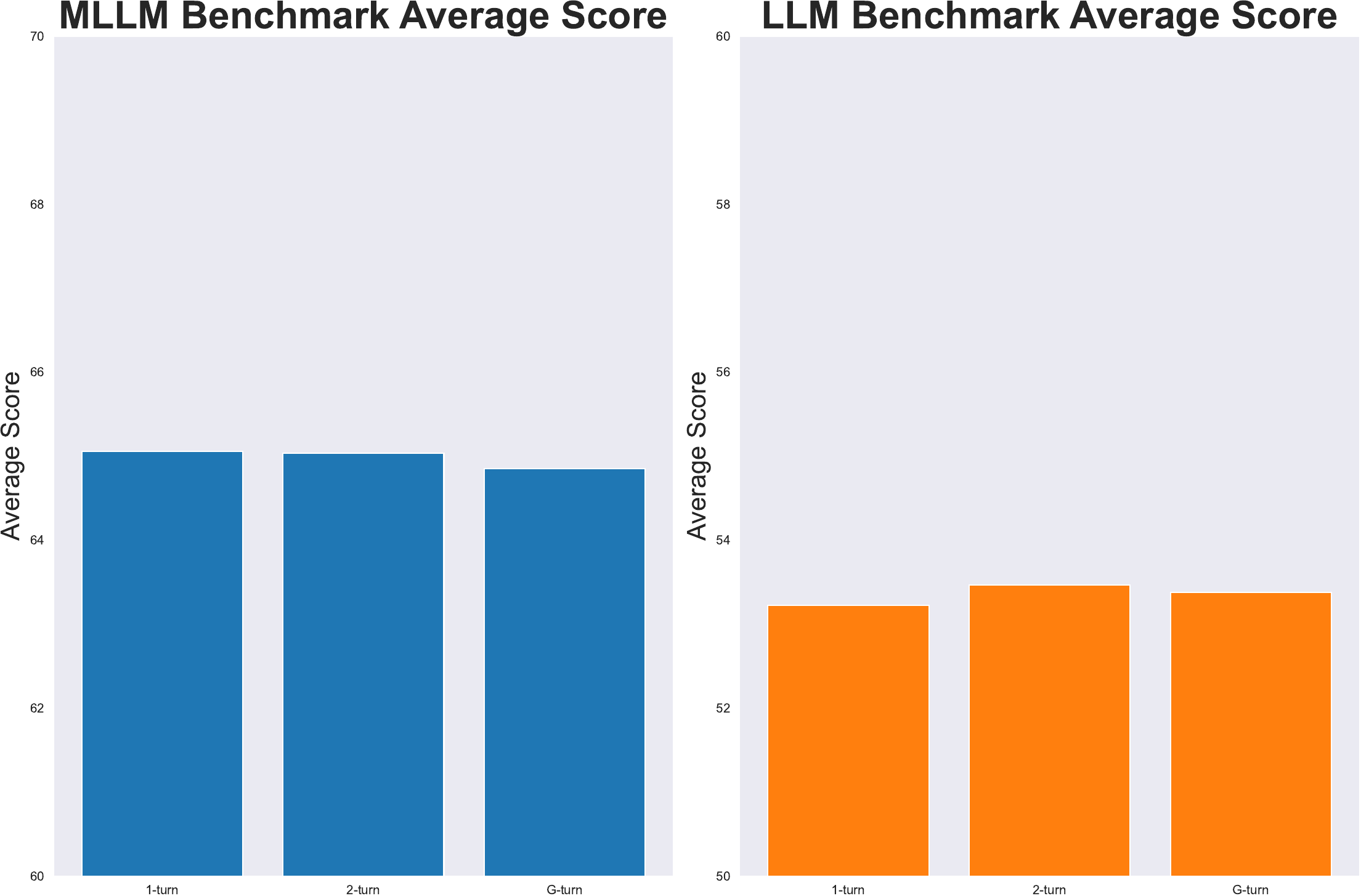}
  \caption{Number of turns in TG-SFT decompressed Data}
  \label{fig:num_turns}
\end{figure}


We investigate the effects of different numbers of conversation turns in TG-SFT generated data, as shown in Figure \ref{fig:num_turns}. We experimented with corpora consisting of all 1-turn, all 2-turn, and a mixture of 1, 2, 3, and 4 turns (referred to as "G-turn" since the distribution of \# turns follows a Gaussian distribution). Results indicate that training with the 2-turn corpus achieves the best performance in LLM benchmarks compared to the other two configurations. 
This could be attributed to the G-turn corpus being too diverse in context length and the 1-turn corpus being too short, which harms the the model during SFT.




\section{Conclusion \& Future Work}

To address the problem of catastrophic forgetting in LLMs and MLLMs, we designed a novel model-agnostic self-decompression method, \textbf{TG (Tree Generation)}, which decompresses knowledge within LLMs into the training corpus. We introduced its variants: TG-SFT for supervised fine-tuning. By utilizing this decompressed corpus, we mitigate the forgetting problem. Experiment results show that using the synthetic data generated from TG, an LLM can preserve its original knowledge and perform on par to using human generated high quality data. The tree structure in TG enable flexible control of speed and diversity, which enables better control of the generation process. 

At last, TG has the potential to enable use cases such as pre-training and knowledge distillation.  We leave a more general version of TG-PT for post-pretraining in future work. We plan to verify if using a decompressed corpus from a strong model can enhance the performance of a weaker model.

\newpage

\clearpage
\newpage
\section{Limitation}
There are several limitations of our works, mainly focusing on the safety issue of synthetic data, how to synthesize data for post-pretraining, and how to evaluate the quality and diversity of synthetic data.

\subsection{Data Leakage Risk, NSFW Content Exposure, and Inaccurate Information}
The TG-SFT method, while effective in fine-tuning MLLMs, may inadvertently facilitate the extraction of training data. Given the auto-regressive nature of LLMs, there is a risk that the model could generate outputs that closely resemble its training data. This poses a significant concern if the training data contains sensitive information.

In the event that the training data includes Not Safe For Work (NSFW) content, the TG-SFT method might inadvertently generate responses that expose or allude to such content. This not only undermines the ethical standards of AI applications but also raises questions about the responsible use of LLMs.

The synthetic data may include many inaccurate facts. If researchers use such data to train the model, it is possible that model outputs fake information. A fact verifier is needed for data synthesizing.

\subsection{Synthetic data for Post-Pretraining}
In this paper, we mainly focus on synthesizing data for SFT. We introduced a variant of \textbf{TG}, denoted as TG-PT, for post-pretraining in section \ref{expsection:tgpt}. However, a more general version of TG-PT for post-pretraining is needed in the future since post-pretraining requires much more data compared with SFT.

\subsection{Synthetic data for Specific Domain}
In our experiments, we attempted to synthesize data within the domain of mathematics. However, we found quality of the synthesized data is not satisfactory. The LLMs tend to generate math concept, and simple/wrong calculations and derivations. SFT with such data resulted in a noticeable degradation in performance on mathematical benchmarks. The challenge of synthesizing high-quality data for specific domains, is an issue that merits further investigation in future research.

\subsection{Benchmarks for Evaluating the Quality and Diversity of Synthetic Data.}
The evaluation benchmark for measuring the quality and diversity of synthetic text data is rare. \cite{wei2024longform} introduce LongFact, a prompt set of 2,280 fact-seeking prompts requiring long-form responses, but LongFact is dependent on LLMs for their operations. Consequently, the capabilities of the utilized LLM have a direct impact on the quality of the LongFact prompts. More universal and model-agnostic benchmarks are needed to evaluate the quality of synthetic data. Moreover, \cite{shaib2024standardizing} propose measurement of text diversity including compression ratios, self-repetition of long n-grams, Self-BLEU, BERTScore, etc., but the computation time for calculating such scores for large amount of training corpus is unacceptable.

\clearpage
\newpage
\bibliography{custom}

\clearpage
\newpage
\appendix
\section{Appendix}
\label{sec:appendix}

\subsection{Experiment Detail for Figure 1}
\begin{itemize}
    \item Following LLaVA \cite{liu2023llava}, we select the MLLM benchmarks including:
    
    gqa, textvqa\_val, pope, mme, seedbench, mmbench\_cn\_dev, mmbench\_en\_dev, scienceqa\_img, vqav2\_val, vizwiz\_vqa\_val

    \item Following LLaMAPro \cite{wu2024llama}, we select the LLM benchmarks including:
    
    arc\_challenge (25-shot), gsm8k (5-shot), hellaswag (10-shot), mmlu (5-shot), winogrande (5\-shot), truthfulqa (0-shot)

    \item We used the llava \cite{liu2023llava} codebase (https://github.com/haotian-liu/LLaVA) to conduct the experiment. We first trained the visual projector for the LLaMA2-7B-chat model and SFT the LLaMA2-7B-chat with this project. All training data and training configurations strictly followed the LLaVA repo. 
    
\end{itemize}

\label{appendix:figure1}

\end{document}